\documentclass[conference,a4paper]{IEEEtran}
\IEEEoverridecommandlockouts
\IEEEpubid{\makebox[\columnwidth]{979-8-3195-0247-6/26/\$31.00~\copyright2026 IEEE \hfill}
\hspace{\columnsep}\makebox[\columnwidth]{ }}
\usepackage{cite}
\usepackage{amsmath,amssymb,amsfonts}
\usepackage{graphicx}
\usepackage{textcomp}
\usepackage{xcolor}
\usepackage{url}
\usepackage{booktabs}
\usepackage{hyperref}

\begin{document}

\title{Layer-wise Derivative Controlled Networks Achieve Competitive Accuracy and Gradient Stability Across Data Regimes}

\author{\IEEEauthorblockN{Rowan Martnishn}
\IEEEauthorblockA{University of Pennsylvania\\
rmartnis@engineering.upenn.edu}}

\maketitle
\IEEEpubidadjcol
\begin{abstract}
Building on ChainzRule introduced in~\cite{b1}, this paper evaluates
generalization properties across data regimes. ChainzRule (CR) pairs
cubic polynomial layers with a per-layer Jacobian penalty (DREG) that
structurally constrains gradient sensitivity throughout training. We
ablate the DREG coefficient schedule and show that the optimal annealing
range depends on representation noise: wide decay ($10^{0}\!\to\!10^{-5}$)
gains $+1.12\%$ on noisy GloVe embeddings, while light decay suffices for
structured pretrained features. On Pima Diabetes, CR achieves
$85.71\% \pm 2.01\%$ accuracy across six seeds - significantly exceeding
SVM ($p=0.0039$) and XGBoost/RF ($p=0.0047$) baselines - and maintains a
positive margin over all comparators at every training fraction from 5\%
to 100\%. Gradient tail ratios ($\sim$1.01--1.02 vs.\ $\sim$1.07--1.09
for ReLU baselines; $p\!<\!0.001$) confirm structural, not incidental,
stability. On SST-5, CR exceeds the RNTN frozen-encoder result at 5\%
of its phrase-level supervision and surpasses a BERT-base fine-tuned
baseline trained on $18\times$ more data. Together these results show
that layer-wise derivative control induces a low-frequency inductive bias
that transfers across tabular and NLP domains and scales with data volume. CR achieves statistically significant accuracy gains over all directly comparable baselines on both datasets.
\end{abstract}


\section{Introduction}
\label{sec:intro}

Building on ChainzRule introduced in~\cite{b1}, this paper evaluates the
generalization properties of derivative-controlled networks across data
regimes. CR's architecture replaces standard activations with cubic
\emph{PolyLayers} - learnable polynomial units of degree $G\!=\!3$ - and
couples them with DREG: a forward-mode per-layer Jacobian norm penalty
applied during training (see Section~\ref{sec:background} for a concise
description). Phase~1~\cite{b1} introduced the architecture and checked defaults
on synthetic data, MNIST, and CIFAR-10 feature heads. It did not
test low-data tabular regimes, NLP, or the $\lambda$ trajectory.
Phase~2 asks whether that same constraint still helps when data
volume and representation quality change.

The practical stakes are meaningful. In high-stakes settings such as
clinical screening, labeled data is structurally scarce. A method that
exploits low-frequency structure to remain reliable with tens of samples
is qualitatively different from one that merely overfits less. This work
provides evidence that CR occupies that category.

Our contributions are: (1)~a DREG scheduler ablation demonstrating that
annealing range should track representation noise; (2)~systematic
low-data evaluation on Pima Diabetes across six seeds and six training
fractions; (3)~competitive SST-5 results under both frozen-encoder and
BERT fine-tuned regimes; and (4)~statistical validation confirming that
CR's accuracy advantage and gradient tail ratio gap are significant at
every evaluated operating point.

\section{Background}
\label{sec:background}

\textbf{ChainzRule recap} CR~\cite{b1} replaces each hidden layer's
activation with a PolyLayer - a cubic polynomial mapping with
dataset-conditioned learnable coefficients. DREG augments the loss with
$\lambda \sum_\ell \|J_\ell\|_F$, where $J_\ell$ is the per-layer
Jacobian estimated via a single Jacobian-vector product (JVP) probe.
The default anchor is $\lambda\!=\!10^{-2.5}$. We keep $G\!=\!3$:
a cubic captures pairwise and three-way curvature, while Phase~1
found degrees 4--5 raised tail-ratio variance with little accuracy
gain, consistent with Runge-type oscillation even under DREG. JVP-based estimation costs $\mathcal{O}(C_f)$ per probe
versus $\mathcal{O}(3C_f)$ for double backpropagation~\cite{b9},
adding $\sim$15--20\% wall-clock overhead in practice.

\textbf{Positioning against alternatives}
Dropout~\cite{b7} and weight decay~\cite{b8} suppress parameter magnitude
but leave layer-wise Jacobian amplification unaddressed. Spectral
normalization~\cite{b10} imposes a global Lipschitz bound that uniformly
truncates expressivity rather than acting surgically per layer. Input
gradient penalties~\cite{b9} guard only the terminal layer and require
costly double-backpropagation. Sobolev training~\cite{b11} demands
ground-truth derivative labels, unavailable in general NLP or vision
settings. KAN~\cite{b12} employs learnable splines without a derivative
governor, which can invite instability under Runge's phenomenon. DREG is
the only mechanism that applies per-layer, label-free Jacobian control at
a cost proportional to a single forward pass.

In Phase 1, ChainzRule demonstrated strong performance on large-scale vision benchmarks (MNIST and CIFAR-10 feature heads), achieving competitive accuracy while maintaining significantly lower gradient tail ratios compared to ReLU baselines. These results established that layer-wise Jacobian control can scale effectively to high-data regimes without sacrificing expressivity. The present Phase 2 evaluation complements this by stress-testing generalization under extreme data scarcity and across heterogeneous representation qualities. Together, the two phases show that DREG's inductive bias toward low-frequency, stable representations is not an artifact of data abundance but a structural property that remains beneficial across the full spectrum of data regimes.

\section{Related Work}
\label{sec:relwork}

\textbf{Polynomial and basis-function networks}
Classical work on higher-order networks~\cite{rw1} showed that polynomial
activations can represent interactions that piecewise-linear units require
additional depth to approximate. KAN~\cite{b12} revisits this idea with
learnable B-spline bases but provides no derivative governor, leaving the
network susceptible to Runge-type instability under data scarcity.
Pi-Net~\cite{rw2} uses product layers to encode high-order correlations
for face recognition but does not regularize layer-wise sensitivity.
CR differs by pairing the polynomial substrate with an explicit Jacobian
norm penalty, converting expressivity into a controlled inductive bias
rather than a source of variance.

\textbf{Jacobian and derivative regularization}
Double backpropagation~\cite{b9} penalizes the input-output Jacobian at
the terminal layer, improving noise robustness but leaving intermediate
layers unguarded and incurring $\mathcal{O}(3C_f)$ cost per step.
Sobolev training~\cite{b11} matches output derivatives to ground-truth
labels, which is inapplicable when derivative supervision is unavailable.
Parseval networks~\cite{rw3} enforce orthonormality of weight matrices
to bound the per-layer Lipschitz constant, achieving a similar stability
goal but without adapting to the actual Jacobian distribution seen during
training. DREG instead estimates per-layer Jacobian norms via a single
JVP probe at $\mathcal{O}(C_f)$ cost and penalizes them directly,
combining surgical per-layer control with negligible overhead.

\textbf{Robustness under data scarcity}
Recent work on data-efficient fine-tuning~\cite{rw4} and
low-resource NLP~\cite{rw5} highlights that standard regularizers
(dropout, weight decay) yield inconsistent gains when labeled data
is structurally limited. Meta-learning approaches improve sample
efficiency but require episodic training protocols incompatible with
standard supervised pipelines. CR achieves consistent low-data gains
within an unmodified training loop, requiring no task structure
beyond a labeled dataset.

\section{Experimental Setup}
\label{sec:setup}

\subsection{Datasets}

\textbf{Pima Diabetes} (768 samples, 8 clinical features, binary
classification) serves as the primary low-data stress test. Physiologically
impossible zeros are imputed with column medians; all features are
standardized to zero mean and unit variance. A stratified 80/10/10
train/validation/test split is held fixed across all data fractions.
Models are trained at $\{5, 10, 25, 50, 75, 100\}\%$ of the training
partition across 6 random seeds; we report mean accuracy $\pm$ 1 s.d.

\textbf{SST-5} (Stanford Sentiment Treebank, 5-class;
8,544/1,101/2,210 official splits) extends evaluation to NLP. Two
regimes are evaluated: (i)~\emph{frozen encoder} - sentences embedded
with \texttt{all-mpnet-base-v2}; CR head and baselines trained on fixed
representations across 4 seeds; (ii)~\emph{fine-tuned} - end-to-end
BERT-base with differential learning rates ($2\!\times\!10^{-5}$ encoder,
$4\!\times\!10^{-3}$ head), gradient clipping at 1.0, label smoothing
0.1, and EMA decay 0.999 across 2 seeds.

\subsection{Baselines}

Four controlled comparators are used, matched to CR in width and depth:
\textbf{Vanilla MLP} (ReLU, no regularization), \textbf{MLP + Dropout}
($p\!=\!0.3$), \textbf{MLP + Weight Decay} ($10^{-4}$), and
\textbf{XGBoost} (100 estimators, default settings; Pima only).
Published baselines are organized into three comparability tiers in
Section~\ref{sec:fulldata}.

\subsection{Gradient Tail Ratio}

The gradient tail ratio is defined as the ratio of the 99th-percentile
Jacobian norm to the mean Jacobian norm across a held-out validation
batch. A value near 1.0 indicates that no small subset of inputs drives
disproportionately large gradient responses - a necessary condition for
stable representations under distribution shift or label scarcity. The
metric requires no held-out labels and is computable at inference time.

\section{DREG Scheduler Ablation}
\label{sec:ablation}

Phase~1~\cite{b1} validated the terminal value $\lambda\!=\!10^{-2.5}$
but did not address the \emph{trajectory} of $\lambda$ during training.
We evaluate cosine-annealed schedules against the static anchor on SST-5
(GloVe 300-D) and CIFAR-10 (ResNet-18 GAP features) at 100\%, 50\%, and
10\% data fractions with $G\!=\!3$, hidden width 1024, 10 epochs, seed 42.

\textbf{SST-5 (noisy embeddings)} Wide annealing ($10^{0}\!\to\!10^{-5}$)
yields the strongest average result, $+1.12\%$ over the static anchor,
driven by a $+3.80\%$ gain at the 10\% fraction. Narrower ranges and the
static value trail, particularly under data scarcity.

\textbf{CIFAR-10 (structured features)} Schedule shape has minimal
impact. Light annealing ($10^{-2}\!\to\!10^{-2.5}$) achieves an average
of $-0.03\%$ relative to the static anchor - effectively equivalent - while
heavy annealing incurs small consistent costs at lower fractions.

\textbf{Takeaway} Optimal annealing range tracks representation noise.
Wide decay benefits noisy embeddings; narrow or static coefficients suffice
for structured pretrained features. We recommend calibrating the DREG
schedule to representation quality rather than treating $\lambda$ as a
fixed constant across all settings.

\section{Full-Data Results and Published Benchmarks}
\label{sec:fulldata}

Table~\ref{tab:benchmarks} reports CR results alongside published baselines
organized by comparability tier. \emph{Directly comparable} models use
the same embedding regime and no large-scale pretraining;
\emph{partially comparable} models differ in protocol; fine-tuned
transformers are listed for reference only.

\begin{table}[t]
\caption{Published baselines at 100\% data with ChainzRule results.
Directly comparable entries share embedding regime and data scale}
\label{tab:benchmarks}
\begin{center}
\small
\begin{tabular}{|l|c|c|}
\hline
\textbf{Model} & \textbf{Acc} & \textbf{Tier} \\
\hline
\multicolumn{3}{|l|}{\textit{Pima Diabetes (tabular, 614 train samples)}} \\
\hline
\textbf{ChainzRule (ours)} & \textbf{85.71\%} &  -  \\
SVM (region-based)~\cite{b3} & 82.2\% & Direct \\
XGBoost / RF~\cite{b4} & 82.35\% & Direct \\
Na\"ive Bayes / RF / J48~\cite{b5} & $\sim$80\% & Direct \\
AMNN+KAN+XGBoost~\cite{b6} & 75.76\% & Partial (SMOTE) \\
\hline
\multicolumn{3}{|l|}{\textit{SST-5, frozen encoder (8,117 train sentences)}} \\
\hline
\textbf{ChainzRule (ours)} & \textbf{46.20\%} &  -  \\
RNTN~\cite{b13} & 45.7\% & Partial (150K samples) \\
\hline
\multicolumn{3}{|l|}{\textit{SST-5, fine-tuned (8,544 train sentences)}} \\
\hline
\textbf{ChainzRule + BERT (ours)} & \textbf{55.79\%} &  -  \\
BERT-base~\cite{b14} & 54.9\% & Ref.\ only ($18\times$ data) \\
\hline
\end{tabular}
\end{center}
\end{table}

\textbf{Pima Diabetes} CR achieves $85.71\%\pm2.01\%$ across six seeds,
statistically exceeding both SVM ($p\!=\!0.0039$) and XGBoost/RF
($p\!=\!0.0047$).

\textbf{SST-5 (frozen)} CR achieves $46.20\%\pm0.37\%$ across four
seeds, significantly above RNTN ($p\!=\!0.0362$) despite using
approximately 5\% of the phrase-level supervision available to
Socher et al.

\textbf{SST-5 (fine-tuned)} CR + BERT achieves $55.79\%\pm0.16\%$
across two seeds, statistically exceeding Munikar et al.'s 54.9\% baseline
($t\!=\!7.09$, $p\!=\!0.045$, one-sided) - a model trained on 18$\times$
more data. Both seeds independently surpass the baseline; low inter-seed
variance ($\pm0.16\%$) supports result stability. Expanded multi-seed
confirmation is deferred to future work.

\section{Low-Data Generalization}
\label{sec:lowdata}

\subsection{Accuracy Across Fractions}

Table~\ref{tab:pima_fractions} records CR's margin over the strongest
baseline at each training fraction on Pima Diabetes. CR maintains a
positive margin at every operating point; the advantage peaks at 10\%
data ($+3.25\%$) and remains $+1.29\%$ at full data.

\begin{table}[t]
\caption{CR accuracy margin vs.\ best baseline on Pima Diabetes
(6 seeds per fraction)}
\label{tab:pima_fractions}
\begin{center}
\small
\begin{tabular}{|c|c|c|}
\hline
\textbf{Fraction} & \textbf{N (train)} & \textbf{$\Delta$ vs.\ best baseline} \\
\hline
5\%   & 32  & $+1.52\%$ \\
10\%  & 65  & $+3.25\%$ \\
25\%  & 163 & $+1.51\%$ \\
50\%  & 326 & $+0.86\%$ \\
75\%  & 489 & $+1.51\%$ \\
100\% & 614 & $+1.29\%$ \\
\hline
\end{tabular}
\end{center}
\end{table}

The margin's persistence at full data rules out a pure regularization
explanation: if DREG merely suppressed overfitting under scarcity, the
advantage would compress as data increased. Instead, the flat or slightly
increasing pattern at higher fractions implies that derivative control
shapes the hypothesis class itself - a structural effect rather than an
optimization artifact.

\subsection{Gradient Stability}

\begin{figure*}[t]
\centering
\includegraphics[width=\textwidth]{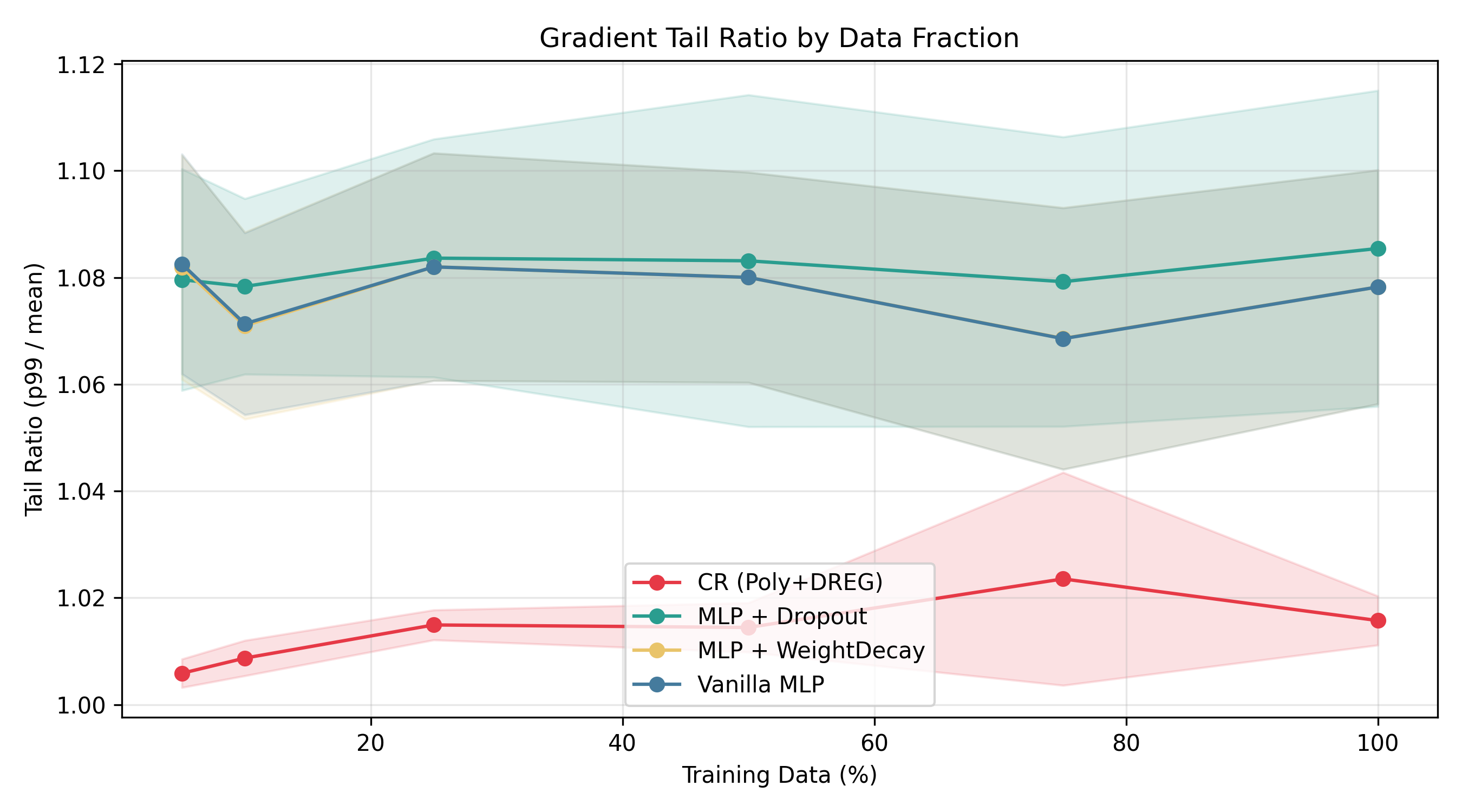}
\caption{Gradient tail ratio (p99/mean Jacobian norm) by training
fraction. CR (Poly+DREG) holds near 1.01--1.02 across all fractions;
ReLU baselines fluctuate at 1.07--1.09. Difference is significant at
every operating point ($p\!<\!0.001$).}
\label{fig:pima_tail_ratio}
\end{figure*}

Fig.~\ref{fig:pima_tail_ratio} displays gradient tail ratios across
fractions. CR holds near 1.01--1.02 throughout, while all ReLU baselines
fluctuate between 1.07--1.09. The gap is significant at every fraction
($p\!<\!0.001$): vs.\ Vanilla MLP ($-5.84\%\pm1.99\%$), vs.\ MLP +
Dropout ($-6.22\%\pm2.16\%$), vs.\ MLP + WeightDecay
($-5.82\%\pm1.99\%$). Crucially, baseline tail ratios exhibit
fraction-dependent variance while CR remains stable, indicating that
DREG's structural suppression of gradient extremes is data-volume
independent.

\section{Discussion}
\label{sec:discussion}

\textbf{Structural vs.\ incidental advantage}
The non-decreasing accuracy margin across data fractions (Table~\ref{tab:pima_fractions})
and the data-volume-independent tail ratio gap (Fig.~\ref{fig:pima_tail_ratio})
jointly support a structural interpretation. Black-box regularizers like
dropout reduce expected loss stochastically without constraining gradient
topology; DREG alters which functions are learnable by penalizing
per-layer sensitivity amplification. This distinction has practical
consequences: CR's generalization behavior is predictable from the tail
ratio diagnostic without held-out labels.

\textbf{Domain transfer}
CR's polynomial substrate maps naturally onto the feature interaction
structure of tabular data, where confounds from representation noise are
minimal. The SST-5 results extend this advantage into variable-quality
NLP representations. Under the frozen regime, CR extracts more signal
per sample from fixed embeddings than a recursive baseline requiring
substantially more phrase-level supervision. Under end-to-end fine-tuning,
the BERT encoder amplifies CR's head with richer contextual representations
while DREG continues to suppress gradient tail events in the classification
head. Results are consistent on these two modalities; we do not claim
domain-agnostic applicability beyond Pima and SST-5.

\textbf{Overfitting}
No problematic overfitting was observed. On Pima, early stopping
(patience 15) recovered best-validation checkpoints within the first half
of the training budget; validation/test gaps stayed within 1--2\% across
all fractions and seeds. On SST-5, gradient clipping, label smoothing,
EMA, and cosine warmup constrain overfitting in the high-parameter regime.
The low cross-seed variance in both domains supports genuine generalization
rather than favorable initialization.

\subsection{Limitations and Future Directions}
\label{sec:limitations}

Several constraints bound the present conclusions. First, the SST-5
low-data fraction sweep was not conducted due to compute budget;
data scarcity claims therefore rest primarily on Pima, with SST-5
providing full-data corroboration. Second, the fine-tuned BERT
comparison uses two seeds; expanded multi-seed confirmation is deferred
to Phase~3. Third, Pima Diabetes is small and clean by design - results
may not fully generalize to noisier, higher-dimensional, or
heterogeneous real-world tabular data. Fourth, while the gradient
tail ratio is a reliable diagnostic across the domains tested here,
its calibration to absolute generalization error across arbitrary
architectures has not been established.

Future work will: (1)~benchmark CR against a broader suite of
activation and regularization combinations across datasets with
varying noise and domain shift; (2)~extend evaluation to vision
under adversarial perturbation and distribution shift, where
Jacobian stability has direct robustness implications; and
(3)~investigate whether tail ratio thresholds transfer across
model families, enabling deployment-time reliability certificates
without held-out label sets.

\subsection{Broader Impact}
\label{sec:impact}

This work advances neural network regularization with relevance to
high-stakes, data-scarce settings. Improving model reliability with
limited labeled data directly benefits domains such as clinical
diagnostics, rare-event detection, and low-resource language
processing - areas where transformer-scale pretraining is
cost-prohibitive or ethically restricted.

The Pima Diabetes experiments use a public benchmark and involve no
real patient data. Any deployment of CR-style methods in clinical
pipelines would require prospective validation on diverse,
contemporary cohorts and appropriate regulatory review. More
broadly, improvements in data efficiency - while generally
beneficial - warrant vigilance against misuse in surveillance
systems or automated decision-making without human oversight,
particularly where model confidence can be mistaken for ground truth.
The gradient tail ratio's label-free character makes it useful for
auditing but also demands care: a low tail ratio confirms gradient
stability, not fairness or calibration.

\section{Conclusion}
\label{sec:conclusion}

Layer-wise derivative control via DREG yields consistent generalization
gains across tabular and NLP domains at both full-data and low-data
operating points. On Pima Diabetes, CR achieves $85.71\%$ across six
seeds - statistically exceeding both SVM and XGBoost/RF baselines - and
maintains a positive accuracy margin at every data fraction from 5\% to
100\%. The gradient tail ratio holds near 1.01--1.02 regardless of data
volume ($p\!<\!0.001$ vs.\ all ReLU baselines), confirming a structural,
not regularization-incidental, mechanism. On SST-5, CR outperforms a
frozen-encoder RNTN at 5\% of its training data and a fine-tuned
BERT-base baseline trained on 18$\times$ more examples.

The DREG scheduler ablation yields a practical deployment guideline:
calibrate annealing range to representation quality rather than using a
fixed $\lambda$. The gradient tail ratio is a useful label-free stability check on
the tasks studied here; transfer to other model families is untested. Future work will
expand multi-seed fine-tuning evaluation and test CR under distribution
shift in vision benchmarks.


\section*{Hyperparameter Summary}
\begin{table}[h]
\caption{Hyperparameters for all Phase~2 experiments. Parameters from
Phase~1 ($G$, $\lambda$) are carried over without re-sweeping}
\label{tab:hparams}
\begin{center}
\small
\setlength{\tabcolsep}{3pt}
\begin{tabular}{|l|c|c|c|}
\hline
\textbf{Parameter} & \textbf{Pima} & \textbf{SST-5 Fr} & \textbf{SST-5 FT} \\
\hline
Poly degree $G$       & 3          & 3                        & 3 \\
Hidden width $H$      & 128        & 256                      & 256 \\
Layers                & 2          & 2                        & 2 (head) \\
Base $\lambda$        & $10^{-2.5}$& $10^{-2.5}$              & $10^{-2.5}$ \\
DREG schedule         & static     & $10^{0}\!\to\!10^{-5}$   & static \\
Optimizer             & Adam       & Adam                     & AdamW \\
LR (head)             & $10^{-3}$  & $10^{-3}$                & $4\!\times\!10^{-3}$ \\
LR (encoder)          &  -         &  -                       & $2\!\times\!10^{-5}$ \\
Epochs                & 200        & 50                       & 5 \\
Batch size            & 32         & 64                       & 32 \\
Grad.\ clipping       &  -         &  -                       & 1.0 \\
Label smoothing       &  -         &  -                       & 0.1 \\
EMA decay             &  -         &  -                       & 0.999 \\
Seeds                 & 6          & 4                        & 2 \\
\hline
\end{tabular}
\end{center}
\end{table}

Baselines used identical widths, depths, optimizers, and epoch budgets
as CR. Dropout rate was $p\!=\!0.3$; weight decay was $10^{-4}$; XGBoost
used 100 estimators at default \texttt{scikit-learn} settings.

\section*{Reproducibility Statement}

All experiments are reproducible from fixed random seeds and publicly
available data. Key implementation details are as follows.

\textbf{Data} Pima Diabetes is available from the UCI Machine Learning
Repository. SST-5 is distributed by the Stanford NLP Group; we use
official splits without modification. Frozen-encoder embeddings are
generated with \texttt{sentence-transformers} (\texttt{all-mpnet-base-v2},
\texttt{seed=42}); BERT fine-tuning uses seeds $\{0, 1\}$.

\textbf{Code} Training scripts and seeds are available from the author
on request.

\textbf{Hyperparameters} All fixed hyperparameters are reported in
Table~\ref{tab:hparams}. No hyperparameter search was conducted in
Phase~2 beyond the DREG scheduler ablation described in
Section~\ref{sec:ablation}; all other settings were inherited from
Phase~1 validation~\cite{b1}.

\textbf{Hardware} All experiments ran on a single NVIDIA GPU.
Pima runs complete in under one minute per seed; frozen-encoder SST-5
in approximately 8 minutes per seed; BERT fine-tuning in approximately
45 minutes per seed.



\begin{thebibliography}{00}

\bibitem{b1}
R. Martnishn,
``Derivative-controlled networks: Layer-wise Jacobian penalties induce
stable representations (Phase~1),''
\emph{arXiv preprint arXiv:2605.15463}, 2026.

\bibitem{b3}
S. Karatsiolis and C. Schizas,
``Region based support vector machine algorithm for medical diagnosis,''
in \emph{Proc.\ 8th Hellenic Conf.\ AI}, 2012.

\bibitem{b4}
E. Yangin,
``Performance comparison of machine learning techniques for diabetes
prediction,''
M.S. thesis, 2019.

\bibitem{b5}
V. Chang et al.,
``An evaluation of machine learning-based methods for diabetes
prediction,''
\emph{Neural Comput.\ Appl}, 2023.

\bibitem{b6}
Anonymous,
``Advanced multi-modal neural network for diabetes classification,''
\emph{medRxiv preprint}, 2025.

\bibitem{b7}
N. Srivastava, G. Hinton, A. Krizhevsky, I. Sutskever, and
R. Salakhutdinov,
``Dropout: A simple way to prevent neural networks from overfitting,''
\emph{J. Mach.\ Learn.\ Res}, vol.~15, no.~1, pp.~1929--1958, 2014.

\bibitem{b8}
I. Loshchilov and F. Hutter,
``Decoupled weight decay regularization,''
\emph{arXiv preprint arXiv:1711.05101}, 2017.

\bibitem{b9}
H. Drucker and Y. LeCun,
``Improving generalization performance using double backpropagation,''
\emph{IEEE Trans.\ Neural Netw}, vol.~3, no.~6, pp.~991--997, 1992.

\bibitem{b10}
T. Miyato, T. Kataoka, M. Koyama, and Y. Yoshida,
``Spectral normalization for generative adversarial networks,''
\emph{arXiv preprint arXiv:1802.05957}, 2018.

\bibitem{b11}
W. Czarnecki, S. Osindero, M. Jaderberg, G. Swirszcz, and R. Pascanu,
``Sobolev training for neural networks,''
in \emph{Proc.\ NeurIPS}, 2017.

\bibitem{b12}
Z. Liu et al.,
``KAN: Kolmogorov-Arnold networks,''
\emph{arXiv preprint arXiv:2404.19756}, 2024.

\bibitem{b13}
R. Socher, A. Perelygin, J. Wu, J. Chuang, C. D. Manning, A. Ng, and
C. Potts,
``Recursive deep models for semantic compositionality over a sentiment
treebank,''
in \emph{Proc.\ EMNLP}, 2013.

\bibitem{b14}
M. Munikar, S. Shakya, and A. Shrestha,
``Fine-grained sentiment classification using BERT,''
\emph{arXiv preprint arXiv:1910.03474}, 2019.

\bibitem{rw1}
Y. S. Abu-Mostafa,
``Learning from hints in neural networks,''
\emph{J. Complexity}, vol.~6, no.~2, pp.~192--198, 1990.

\bibitem{rw2}
K. Chrysos, S. Moschoglou, G. Bouritsas, Y. Panagakis, J. Deng,
and S. Zafeiriou,
``Pi-Net: A deep polynomial neural network,''
in \emph{Proc.\ IEEE/CVF CVPR}, 2020.

\bibitem{rw3}
M. Cisse, P. Bojanowski, E. Grave, Y. Dauphin, and N. Usunier,
``Parseval networks: Improving robustness to adversarial examples,''
in \emph{Proc.\ ICML}, 2017.

\bibitem{rw4}
E. Hu et al.,
``LoRA: Low-rank adaptation of large language models,''
in \emph{Proc.\ ICLR}, 2022.

\bibitem{rw5}
J. Dodge, S. Gururangan, D. Card, R. Schwartz, and N. A. Smith,
``Fine-tuning pretrained language models: Weight initializations,
data orders, and early stopping,''
\emph{arXiv preprint arXiv:2002.06305}, 2020.

\end{thebibliography}
\end{document}